\documentclass[10pt,twocolumn]{article} 
\usepackage{simpleConference}
\usepackage{times}
\usepackage{graphicx}
\usepackage{amssymb}
\usepackage{color}
\usepackage{amsmath}
\usepackage{url,hyperref}
\usepackage{authblk}
\usepackage{booktabs}
\usepackage{multirow}
\usepackage{caption}
\usepackage{float}
\usepackage{setspace}
\begin{document}

\title{Lightweight Vision Transformer with Cross Feature Attention}
\author[1]{Youpeng Zhao}
\author[1,2]{Huadong Tang}
\author[1]{Yingying Jiang}
\author[1]{Yong A}
\author[2]{Qiang Wu}
\affil[1]{Samsung Research China, Beijing}
\affil[2]{School of Electrical and Data Engineering, University of Technology Sydney}
\maketitle
\begin{abstract}
\noindent \textit{Recent advances in vision transformers (ViTs) have achieved great performance in visual recognition tasks. Convolutional neural networks (CNNs) exploit spatial inductive bias to learn visual representations, but these networks are spatially local. ViTs can learn global representations with their self-attention mechanism, but they are usually heavy-weight and unsuitable for mobile devices. In this paper, we propose \textit{cross feature attention} (XFA) to bring down computation cost for transformers, and combine efficient mobile CNNs to form a novel efficient light-weight CNN-ViT hybrid model, XFormer, which can serve as a general-purpose backbone to learn both global and local representation. Experimental results show that XFormer outperforms numerous CNN and ViT-based models across different tasks and datasets. On ImageNet-1K dataset, XFormer achieves top-1 accuracy of 78.5\% with 5.5 million parameters, which is 2.2\% and 6.3\% more accurate than EfficientNet-B0 (CNN-based) and DeiT (ViT-based) for similar number of parameters. Our model also performs well when transferring to object detection and semantic segmentation tasks. On MS COCO dataset, XFormer exceeds MobileNetV2 by 10.5 AP ($22.7 \to 33.2$ AP) in YOLOv3 framework with only 6.3M parameters and 3.8G FLOPs. On Cityscapes dataset, with only a simple all-MLP decoder, XFormer achieves mIoU of 78.5 and FPS of 15.3, surpassing state-of-the-art lightweight segmentation networks.}
\end{abstract}

\section{Introduction}
\noindent Modern computer vision has witnessed great success from deep learning thanks to the emergence of convolutions neural networks (CNNs) \cite{lenet,alexnet,vgg,resnet} and large-scale GPU computing \cite{gpu1,gpu2}. CNNs focus on processing spatially local information to learn the visual representation and have become the standard approach towards many vision tasks. Vision transformers (ViTs) recently emerged as an alternative towards representation learning and have demonstrated its ubiquitous performance for a wide variety of visual recognition tasks, such as image classification \cite{vit,deit,swin}, semantic segmentation \cite{setr,segmentor,Segformer,lvt} and object detection \cite{yolos,vitdet}. ViTs treat images as sequences of patches, and split them into image tokens to model their long range interaction using self-attention \cite{attention}. Compared to CNNs, ViTs have the advantages of global processing ability and exhibit excellent scalability with the further growing dataset size, e.g., JFT-300M \cite{jft}.\\[6pt]
However, in the regime of \textit{lightweight} and mobile networks, MobileNets \cite{mbv,mbv2,mbv3} and EfficientNets \cite{efficientnet} still dominate over ViTs. For instance, under a parameter budget of 6 million, DeiT \cite{deit} is 3\% less accurate than MobileNetV3 \cite{mbv3} on ImageNet classification task and has $6 \times$ more FLOPs. Light-weight CNNs have served as a general-purpose backbone for many mobile vision tasks, while most current ViT-based models are heavy-weight, hard to optimize and scale down, requiring expensive decoders for downstream tasks. Recent works \cite{Segformer,lvt} have demonstrated the possibility of designing lightweight ViT models for specific tasks like semantic segmentation, but ViT-based models are far from being considered a practical option for mobile devices. Therefore, it is of crucial importance to design an efficient and lightweight ViT model that can generalize well for different tasks. \\[6pt]
Several methods \cite{mobileformer,mobilevit,linformer,sparsevit,reformer} have been proposed to improve ViTs. A straightforward idea is to combine convolutional layers and transformers by using convolution at earlier stages or interwining convolution into each transformer block. These methods \cite{mobileformer,mobilevit} aim to bring down computation cost and make the optimization easier by introducing spatially local bias. Another approach is to optimize the self-attention operation in transformers. The original self-attention has \textbf{quadratic} computation complexity with respect to token number, which causes huge bottleneck for transferring to downstream tasks. Previous works adopt methods such as factorization \cite{sparsevit}, low-rank approximation \cite{linformer} or hashing \cite{reformer} to reduce complexity, but the general trade-off is a notable performance drop.\\[6pt]
In this paper, we first address the efficiency bottleneck in transformers by proposing a novel \textit{cross feature attention} (XFA) module. Instead of performing matrix multiplication via token dimension, XFA handles attention through feature dimension by constructing intermediate context and feature scores, which produces only \textbf{linear} complexity. Within XFA, we also eliminate the softmax operation and replace it with a learnable parameter to dynamically adjust scaling factor for different transformer layers. Based on XFA, we then propose a new light-weight CNN-ViT hybrid model, XFormer, consisting of XFA transformer blocks and efficient CNN blocks, which leverages both the spatial local inductive bias from CNNs and global information from transformers. \\[6pt]
Our method achieves solid performance on three standard visual recognition tasks. On ImageNet-1K dataset, XFormer yields 78.5\% top-1 classification accuracy at 1.7G FLOPs, outperforming MobileNetV3 \cite{mbv3} and Mobile-Former \cite{mobileformer} by a clear margin. We further demonstrate the generalizability of our model as a reliable backbone network. On MS COCO dataset, YOLOv3 \cite{yolov3} with XFormer improves detection results of state-of-the-art CenterNet \cite{centernet} by 7.9 AP ($25.3 \to 33.2$ AP), with 56\% less parameters and 22\% less FLOPs. On Cityscapes dataset, XFormer exceeds the performance of MobileViT \cite{mobilevit} in SegFormer \cite{Segformer} framework, by 2.9 mIoU ($75.6 \to 78.5$ mIoU) with reduced computation cost.\\[6pt]
The main contributions of our paper are as follows: (1) we propose a new self-attention method, \textit{cross feature attention} (XFA), which operates on feature dimension and produces \textbf{linear} complexity with respect to token number; (2) we propose a novel light-weight CNN-ViT hybrid model, called XFormer, based on our proposed XFA module, and validate its superiority on various vision datasets.
\begin{figure*}[!ht]
\begin{center}
  \includegraphics[width=1.0\linewidth]{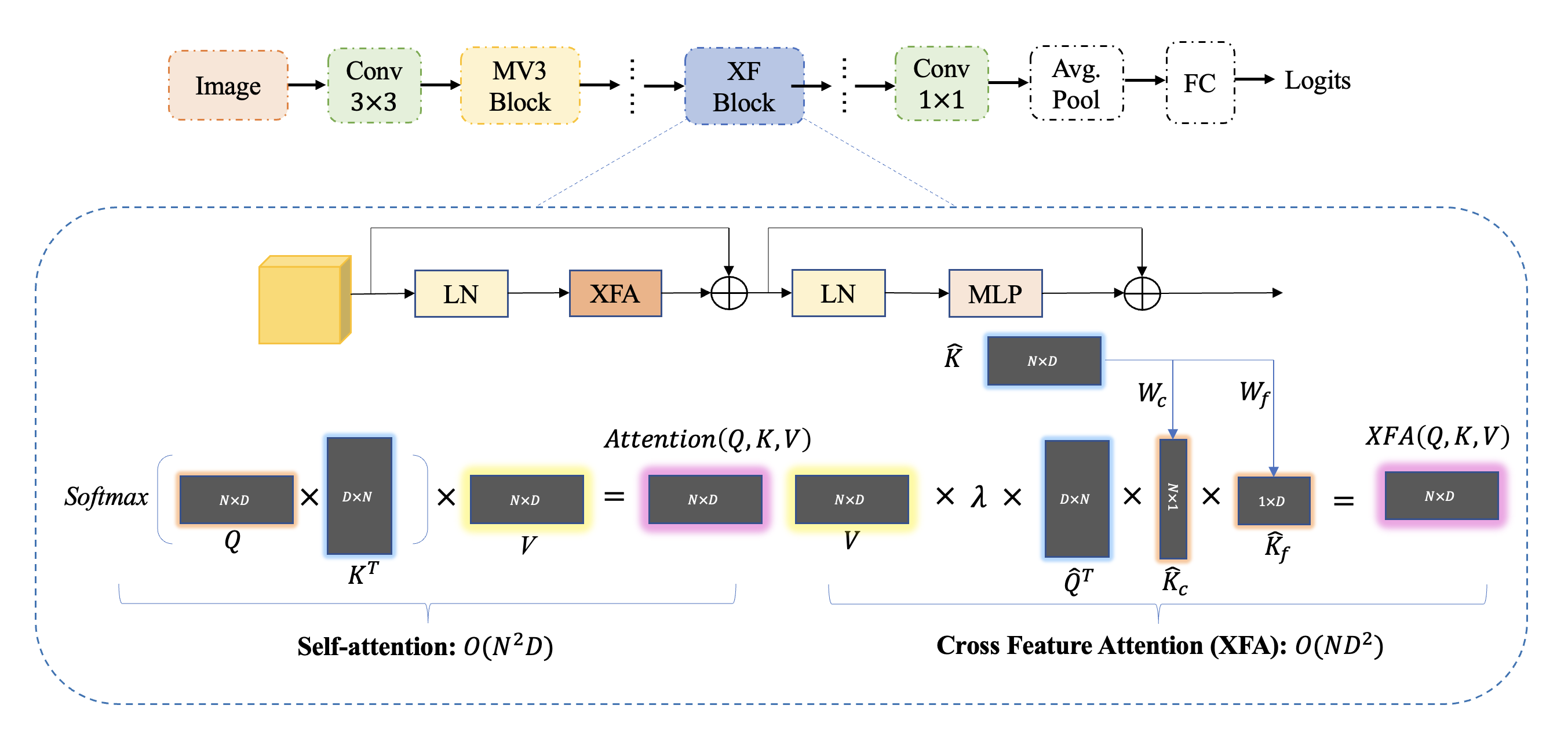}
\end{center}
  \caption{Cross feature transformer (XFormer). The top row is overall hierarchical architecture. The bottom row is XF block and the visual comparison of between our proposed cross feature attention (XFA) and self-attention in ViT \cite{vit}.}
\label{fig:arch}
\end{figure*}

\section{Related Work}
\noindent \textbf{Lightweight CNNs.} 
CNNs \cite{lenet,alexnet,vgg,resnet} are considered standard network model for modern computer vision tasks. General trend to achieve higher accuracy is to design deeper and more complicated networks. To make CNNs suitable for computationally limited platforms such as mobile devices, methods such as pruning, neural architecture search (NAS), knowledge distillation and efficient design of convolutional layer have been proposed. Pruning-based methods \cite{prun1,prun2,prun3} aim to remove redundant weights to compress and accelerate the original network. NAS \cite{darts,proxylessnas,ofa} focuses on efficient strategies to find the optimal architecture under restrictions like FLOPS and inference speed. Knowledge distillation \cite{kd1,kd2,kd3} transfers useful information from a heavy teacher network to a light-weight student network by minimizing the Kullback-Leibler divergence. Other works have focused on improving convolutional layers to make CNNs lightweight and mobile-friendly. MobileNets \cite{mbv,mbv2,mbv3} adopt depthwise and pointwise convolutions to efficiently encode local features. EfficientNets \cite{efficientnet} use compound efficient to scale down network depth/width/resolution. ShuffleNets \cite{shufflev1,shufflev2} utilize group convolution and channel shuffling techniques to further simplify pointwise convolution. AdderNets \cite{addernet1,addernet2} employ additions instead of multiplications to bring down computation cost. \\[6pt]
\noindent \textbf{Vision Transformers (ViTs).} 
Vision Transformer \cite{vit} is the first work to apply NLP transformer \cite{attention} for image recognition tasks and achieves CNN-level accuracy when trained with extremely large scale dataset, e.g., JFT-300M \cite{jft}. DeiT \cite{deit} subsequently explores training strategies on smaller ImageNet-1K dataset. Swin-T \cite{swin} presents a hierarchical ViT structure with shifted local windows for high resolution images. Dynamic ViTs \cite{dynamicvit} cascades multiple transformers to dynamically adjust token number for different input images to improve inference speed. Despite their excellent performances, ViTs have shown substandard optimizablity and are difficult to train. One of the underlying causes is lack of spatial inductive bias \cite{mobileformer}. Afterwards, a series of methods incorporating such biases using convolutions have been proposed. ConvViT \cite{convvit} uses a gated positional self-attention to produce soft convolutional inductive bias. ViT{$_{c}$} \cite{vitc} replaces patchified stem with convolutional stem for better performance. MobileViT \cite{mobilevit} adopts MobileNets-like design to construct lightweight ViTs, while Mobile-Former \cite{mobileformer} explores a bidirectional parallel structure of MobileNets and transformer.\\[6pt]
\noindent \textbf{Self-Attention.} 
Multi-head self-attention (MHSA) \cite{attention} is first introduced to address parallelization issues in training recurrent neural networks (RNNs) for large NLP datasets. Previous works \cite{reformer,linformer,sparsevit,xcit} have tried to address the computational bottleneck by reducing complexity of self-attention in transformer blocks from $\mathcal{O}(N^2)$ to $\mathcal{O}(N\sqrt{N})$ or $\mathcal{O}(N \log{N})$, but they suffer from large performance degradation. Sparse transformers \cite{sparsevit} adopt a factorized operation for self-attention to reduce the complexity to $\mathcal{O}(N\sqrt{N})$, but its effectiveness is only shown on low-resolution dataset. Linformer \cite{linformer} replaces self-attention with low-rank approximation operation, but large batch matrix decomposition is still computationally expensive and the efficiency gain is only noticeable for large token number ($N>2048$). These methods, though achieving significant speed-up over original MHSA, are not suitable for a general-purpose mobile backbone, which is required to be easily transferred to downstream tasks without significant performance drop.

\section{Method}
\noindent A standard ViT model first reshapes the input $X \in \mathbb{R}^{H \times W \times C}$ into a sequence of flattened patches $X_{f} \in \mathbb{R}^{N \times PC}$ using patch size $h \times w$, where $P=hw$ and $N = \frac{HW}{P}$ denotes token number. $X_{f}$ is then projected to a fixed $D$-dimensional space $X_{D} \in \mathbb{R}^{N \times D}$ and using a stack of transformer blocks to learn inter-patch representations. ViTs usually require more parameters to learn visual information, due to their ignoring of spatial inductive bias. Moreover, the expensive computation of self-attention in transformers causes bottleneck in optimizing such models. \\[6pt]
In this section, we propose XFormer, our efficient, light-weight CNN-ViT framework to combat the above problems in ViTs. We first introduce a new method to improve the efficiency of self-attention, and then we illustrate the architecture design of our new CNN-ViT hyrbid model. 
\subsection{Cross Feature Attention (XFA)}
\noindent \textbf{Attention Overview.} 
One of the main computation bottlenecks in transformer lies in the self-attention layer. In the original self-attention process, $X_{D}$ is first used to produce queries $Q$, keys $K$ and values $V$ via linear projections. They all have the same dimension of $ N \times D$, where $N$ is the image token number, each of dimensionality of $D$. Attention score is then calculated as:

\begin{equation}
  \textrm{Attention}(Q,K,V) = \sigma (\frac{QK^{T}}{\sqrt{d_{head}}})V
\end{equation}

\noindent where $\sigma$ is softmax operation and $d_{head}$ is the head dimension. The computation complexity is for calculating attention score is $\mathcal{O}(N^2D)$. The \textbf{quadratic} complexity nature of self-attention induces huge computation bottleneck, which makes it difficult for ViT models to scale down on mobile devices. \\[6pt]
\noindent \textbf{Efficient Attention.} 
To address the quadratic complexity problem in self-attention, we propose a new structure of attention module, called \textit{cross feature attention} (XFA). Following previous work \cite{xcit}, we first apply $L_2$ normalization on queries $Q$ and keys $K$ along feature dimension $D$:
\begin{equation}
  \hat{Q} = \frac{Q}{\| Q \|_{2}^{D}}, \hat{K} = \frac{K}{\| K \|_{2}^{D}}
\end{equation}
\noindent The intuition of $QK^{T}$ in original self-attention is to localize important image patches that should be attended to. But direct computation causes unnecessary redundancy and computation overhead. Instead, we construct two intermediate scores for $K$: query context scores $\hat{K_{c}}$ and query feature scores $\hat{K_{f}}$. We use two convolutional kernel matrices $W_c \in \mathbb{R}^{1 \times N}$ and $W_f \in \mathbb{R}^{1 \times D}$ to calculate $\hat{K_{c}} \in \mathbb{R}^{1 \times D}$ along the token dimension $N$ and $\hat{K_{f}} \in \mathbb{R}^{N \times 1}$ along the feature dimension $D$. With the help of convolutional filters, our intermediate score vectors can represent a more compact representation for calculating attention maps, while also reducing computation cost. $\hat{K_{c}}$ and $\hat{K_{f}}$ are formulated as:
\begin{equation}
  \hat{K_{c}} = W_c\hat{K}
\end{equation}
\begin{equation}
  \hat{K_{f}} = (W_f\hat{K}^{T})^{T}
\end{equation}
Finally, we define our cross feature attention (XFA) as:
\begin{equation}
  \textrm{XFA}(Q,K,V) = V \lambda \hat{Q}^T\hat{K_{f}}\hat{K_{c}}
\end{equation}
\noindent where $\lambda$ is a temperature parameter to dynamically adjust scaling factor in different transformer layers, allowing for more training stability. Noted that normalization bounds the attention values within certain range, so we drop the redundant and expensive softmax operation. Unlike original self-attention, which has quadratic complexity, our module reduce the computation cost from $\mathcal{O}(N^2D)$ to $\mathcal{O}(ND^2)$, which only scales linearly with $N$.\\[6pt]
\noindent \textbf{Comparison with Self-attention.} 
The main differences between our proposed XFA module and original attention are: (1) XFA calculates attention maps along the feature dimension $D$, by constructing intermediate query context and feature scores, which largely reduces the computation cost; (2) XFA uses a learnable temperature scaling parameter to adjust for normalization and is free from softmax operation.\\
Our method provides a fix for the \textbf{quadratic} complexity problem and is more efficient and mobile-friendly for resource-constraint devices. Visual comparison can be seen in Fig. \ref{fig:arch}.
\subsection{Building XFormer}
\begin{table*}[ht!]
\begin{center}
\begin{tabular}{lccccc}
\hline
Input Size & Layer & Out. channels & Repeat & Stride & Stage \\ \hline
$224^2 \times 3$ & Conv2d ($3 \times 3$) $\downarrow 2$ & 16 & 1 & 2 & Stem$_{in}$ \\ \hline
$112^2 \times 16$ & MV3 & 32 & 1 & 1 & 1 \\ \hline
$112^2 \times 32$ & MV3 $\downarrow 2$ & 64 & 1 & 2 & \multirow{2}{*}{2} \\
$56^2 \times 64$ & MV3 & 64 & 2 & 1 &  \\ \hline
$56^2 \times 64$ & MV3 $\downarrow 2$ & 96 & 1 & 2 & \multirow{2}{*}{3} \\
$28^2 \times 96$ & XF Block & 96 & 2 & - &  \\ \hline
$28^2 \times 96$ & MV3 $\downarrow 2$ & 128 & 1 & 2 & \multirow{2}{*}{4} \\
$14^2 \times 128$ & XF Block & 128 & 3 & - &  \\ \hline
$14^2 \times 128$ & MV3 $\downarrow 2$ & 160 & 1 & 2 & \multirow{2}{*}{5} \\
$7^2 \times 160$ & XF Block & 160 & 4 & - &  \\ \hline
$7^2 \times 160$ & Conv2d ($1 \times 1$) & 640 & 1 & 1 & Stem$_{out}$ \\ \hline
$7^2 \times 640$ & AvgPool ($7 \times 7$) & 640 & 1 & - & Global Pooling \\ \hline
$1^2 \times 640$ & Linear & 1000 & 1 & - & Classifier Head \\ \hline
 & \textbf{Network Parameters} &  & 5.5 M  &  &  \\ \hline
\end{tabular}
\end{center}
\caption{\label{tab:model}Model specification of XFormer. MV3 denotes MobileNetV3 block.}
\label{tab:model}
\end{table*}

\begin{table*}[ht!]
\begin{center}
\begin{tabular}{|l|c|cccccccccc|}
\hline
\multirow{3}{*}{Model} & \multirow{3}{*}{\begin{tabular}[c]{@{}c@{}}Attention\\ Module\end{tabular}} & \multicolumn{10}{c|}{Image Resolution} \\ \cline{3-12} 
 &  & \multicolumn{2}{c|}{$256^2$} & \multicolumn{2}{c|}{$512^2$} & \multicolumn{2}{c|}{$768^2$} & \multicolumn{2}{c|}{$1024^2$} & \multicolumn{2}{c|}{$1280^2$} \\ \cline{3-12} 
 &  & Time & \multicolumn{1}{c|}{Mem} & Time & \multicolumn{1}{c|}{Mem} & Time & \multicolumn{1}{c|}{Mem} & Time & \multicolumn{1}{c|}{Mem} & Time & Mem \\ \hline
DeiT & MHSA & 5.02 & \multicolumn{1}{c|}{160} & 10.33 & \multicolumn{1}{c|}{789} & 37.02 & \multicolumn{1}{c|}{2554} & 90.81 & \multicolumn{1}{c|}{6546} & N/A & OOM \\
MobileViT & MHSA & 7.67 & \multicolumn{1}{c|}{426} & 13.38 & \multicolumn{1}{c|}{1856} & 42.96 & \multicolumn{1}{c|}{4965} & 98.06 & \multicolumn{1}{c|}{10837} & 236.9 & 20992 \\
\multirow{2}{*}{XFormer} & MHSA & 9.34 & \multicolumn{1}{c|}{520} & 16.92 & \multicolumn{1}{c|}{2434} & 35.12 & \multicolumn{1}{c|}{7031} & 109.77 & \multicolumn{1}{c|}{16426} & N/A & OOM \\
 & XFA & 10.82 & \multicolumn{1}{c|}{485} & 10.92 & \multicolumn{1}{c|}{1870} & 28.88 & \multicolumn{1}{c|}{4176} & 49.89 & \multicolumn{1}{c|}{7403} & 76.56 & 10551 \\ \hline
\end{tabular}
\end{center}
\caption{\label{tab:memory}Inference time (ms) and GPU memory usage (MB) of various ViT-based models. OOM denotes GPU device out-of-memory. All measurements are performed with a batch size of 2 on a single Titan RTX 24GB GPU.}
\label{tab:memory}
\end{table*}

\noindent \textbf{MobileNetV3 Block.} MobileNetV2 \cite{mbv2} first introduced inverted residual and linear bottleneck to make more efficient layer structures. MobileNetV3 \cite{mbv3} subsequently adds squeeze excitation (SE) \cite{se} module to attend larger representation features. Recent works \cite{convvit,vitc,mobileformer,mobilevit} have demonstrated the legitimacy of boosting ViT performance by incorporating convolutional layers in the early stages of ViTs. Inspired by this intuition, we continue to explore this CNN-ViT hybrid design for our light-weight model. Operations within MobileNetV3 (MV3) block can be formulated as:
\begin{equation}
  x_{out} = x_{in} + \textrm{Conv}_{pw}(\textrm{SE}(\textrm{Conv}_{dw}(x_{in})))
\end{equation}
\noindent where $x_{in}$ is input feature from previous layer, $x_{out}$ is output feature of MV3 Block and SE is the squeeze excitation module. $\textrm{Conv}_{dw}$ denotes depth-wise convolution operation and $\textrm{Conv}_{pw}$ denotes piece-wise convolution operation. \\[6pt]
\noindent \textbf{XF Block.} 
Taking advantages of the proposed cross feature attention module, we introduce XF Block, a carefully designed light-weight transformer module. Operations within XF block can be formulated as:
\begin{equation}
  x_{mid} = x_{in} + \textrm{XFA}(\textrm{LN}(x_{in}))
\end{equation}
\begin{equation}
  x_{out} = x_{mid} + (\textrm{MLP}(\textrm{LN}(x_{mid}))
\end{equation}
\noindent where $x_{in}$ is input feature from previous layer, $x_{mid}$ is feature from XFA module and $x_{out}$ is output feature of XF Block. LN denotes layer normalization \cite{layernorm} operation and MLP denotes fully-connected layer.  \\[6pt]
\noindent \textbf{Patch Size Choice.}  ViT-based models normally employ patch size $8 \times 8$, $16 \times 16$ or even $32 \times 32$ for larger models (ViT-Large \cite{vit}). One of the advantages of having larger patch size lies in that for image-level tasks such as classification, ViT can efficiently extract image patch information without adding too much computation overhead. Recent works \cite{Segformer,mobilevit} have shown that when transferring to downstream tasks such as semantic and instance segmentation, smaller patch size should be favored, since it can enhance transformer's ability to learn better pixel level information, which generally leads to better performance. Moreover, with smaller patch size, token number $N$ is much larger. Our linear-complexity XFA module can avoid the potential computation bottleneck. In our network design, patch size is set to $2 \times 2$ for each XF block. \\[6pt]
\noindent \textbf{XFormer.} 
Building on MobileNetV3 and XF blocks, we propose XFormer, a CNN-ViT hybrid light-weight model, consisting of stacked MobileNetV3 blocks and XF blocks to learn both global and local context information. Like previous works on designing efficient CNNs, our network consists of convolutional stem Stem$_{in}$ block to extract primitive image features, and after features are processed by CNN and transformer blocks, we use Stem$_{out}$, global pooling and fully-connected layers to produce final logit predictions. \\[6pt]
In our main processing blocks, XFormer has five stages. The first two stages contains only MobileNetV3 (MV3) blocks, since convolutional blocks are more capable to extract important image-level feature representation than all-ViT models and subsequently help transformer blocks see better \cite{vitc}. Each of the last three stages includes one MV3 block and a few XF blocks. Combining the local inductive bias from MV3 blocks and global information from transformer blocks, our network can learn a more comprehensive feature representation, which can be easily transferred to different downstream tasks. \\[6pt]
Previous work \cite{sss} on ViTs suggests larger MLP ratio should be used in deeper transformer layers and $Q$-$K$-$V$ dimension should be relatively smaller than embedding dimension to allow for better trade-off between performance and model size. We follow these suggestions and design our light-weight model accordingly. For XF blocks in three different stages, their MLP ratios, embedding dimensions and $Q$-$K$-$V$ dimensions are set to be $(2, 2, 3)$, $(144, 192, 240)$ and $(96, 96, 96)$ respectively. In fully-connected layers, we replace ReLU with GELU \cite{gelu} activations; we use SiLU \cite{SiLU} in all other layers. For MV3 blocks, expansion ratio is all set to 4. The model specification is shown in Table \ref{tab:model}. \\[6pt]
\noindent \textbf{Model efficiency.} 
The total parameter size of our model is only 5.5 million. When compared to ViT-based models of similar size, our model can process high resolution images more efficiently and avoids potential memory bottleneck (see Table \ref{tab:memory}). For instance, when input resolution is $1024 \times 1024$, XFormer demonstrate almost $2 \times$ improvement in inference speed and 32\% reduction in GPU memory usage compared to MobileViT, which uses original self-attention. Our model can easily handle high resolution throughput without memory bottleneck. Above all, XFormer delivers better accuracy than compared models (see Table \ref{tab:imagenet}), achieving great balance between model size and performance.

\begin{table*}[ht!]
\begin{center}
\begin{tabular}{lcccc}
\hline
 Model & Type & Params (M) & Top-1 (\%) & FLOPs (G) \\ \hline
 MobileNetV2 \cite{mbv2} & CNN & 3.5 & 73.3 & 0.3\\ 
 MobileNetV3 \cite{mbv3} & CNN & 5.4 & 75.2 & 0.2\\ 
 ResNet-50 \cite{resnet} & CNN & 25.6 & 76.0 & 4.1\\
 DenseNet \cite{densenet} & CNN & 14.0 & 76.2 & 0.6\\
 EfficientNet-B0 \cite{efficientnet} & CNN & 5.3 & 76.3 & 0.5\\ \hline
 DeiT \cite{deit} & ViT & 5.7 & 72.2 & 1.3\\
 XCiT \cite{xcit} & ViT & 6.7 & 77.1 & 1.6\\
 CrossViT \cite{crossvit} & ViT &8.8 & 77.1 & 2.0\\
 Swin-T \cite{swin} & ViT &7.3 & 77.3 & 1.0\\ \hline
 ConvViT \cite{convvit} & CNN+ViT & 6.0 & 73.1 & 1.0\\ 
 Mobile-Former \cite{mobileformer} & CNN+ViT & 9.4 & 76.7 & 2.1\\ 
 MobileViT \cite{mobilevit} & CNN+ViT & 5.6 & 78.4 & 1.9 \\ 
 \textbf{XFormer} & CNN+ViT & 5.5 & \textbf{78.5} &  1.7 \\ \hline
\end{tabular}
\end{center}
\caption{\label{tab:imagenet}Comparison to state of the art methods on ImageNet classification dataset.}
\end{table*}


\section{Experiments}
\noindent In this section, we evaluate our proposed XFormer on three common vision tasks: image classification, object detection and semantic segmentation.

\subsection{Image Classification}
\noindent \textbf{Dataset.} We use ImageNet-1K, a subset of ILSVRC 2012 \cite{ImageNet} dataset, to train and evaluate our method for image classification. The training and validation set contains 1.28 million and 50K images respectively, with 1000 object classes in total. \\[6pt]
\noindent \textbf{Implementation details.} Our model is trained for 300 epochs with batch size of 1024 using AdamW \cite{adamw} optimizer, and label smoothing cross-entropy loss. Smoothing factor is set to 0.1. We adopt linear learning rate warmup from 0.0002 to 0.002 for 3K iterations, followed by cosine learning rate decay back to 0.0002. Weight decay is set to 0.05 and gradient clipping threshold is set to 10. Random resized cropping, horizontal flipping, random erasing \cite{randomerasing}, RandAug \cite{randaug}, Mixup \cite{mixup} and CutMix \cite{mixup} are applied for data augmentation. During inference, we use exponential moving average (EMA) of model weights. We report classification performance using top-1 (\%) accuracy metrics. \\[6pt]
\noindent \textbf{Comparison with CNNs.} Table \ref{tab:imagenet} shows that XFormer outperforms other light-weight CNN-based models across different model sizes. XFormer improves MobileNetV2 by 5.2\%, MobileNetV3 by 3.3\% and EfficientNet-B0 by 2.2\%. When compared to heavy-weight CNN models, our model produces better top-1 accuracy than ResNet-50 \cite{resnet} and DenseNet \cite{densenet} with much less parameters. \\[6pt]
\noindent \textbf{Comparison with ViTs.} When compared to ViT variants that are trained from scratch on ImageNet-1K dataset, XFormer delivers better performance than DeiT ($+6.3\%$), ConvViT ($+5.4\%$) and CrossViT ($+1.4\%$) for similar or less parameters. Our model also outperforms two state-of-the-art light-weight CNN-ViT hybrid models, Mobile-Former and MobileViT with fewer FLOPS and parameters. This demonstrates that our proposed XFormer is capable of learning representation efficiently.
\begin{figure}[!ht]
 \begin{center}
   \includegraphics[width=1.0\linewidth]{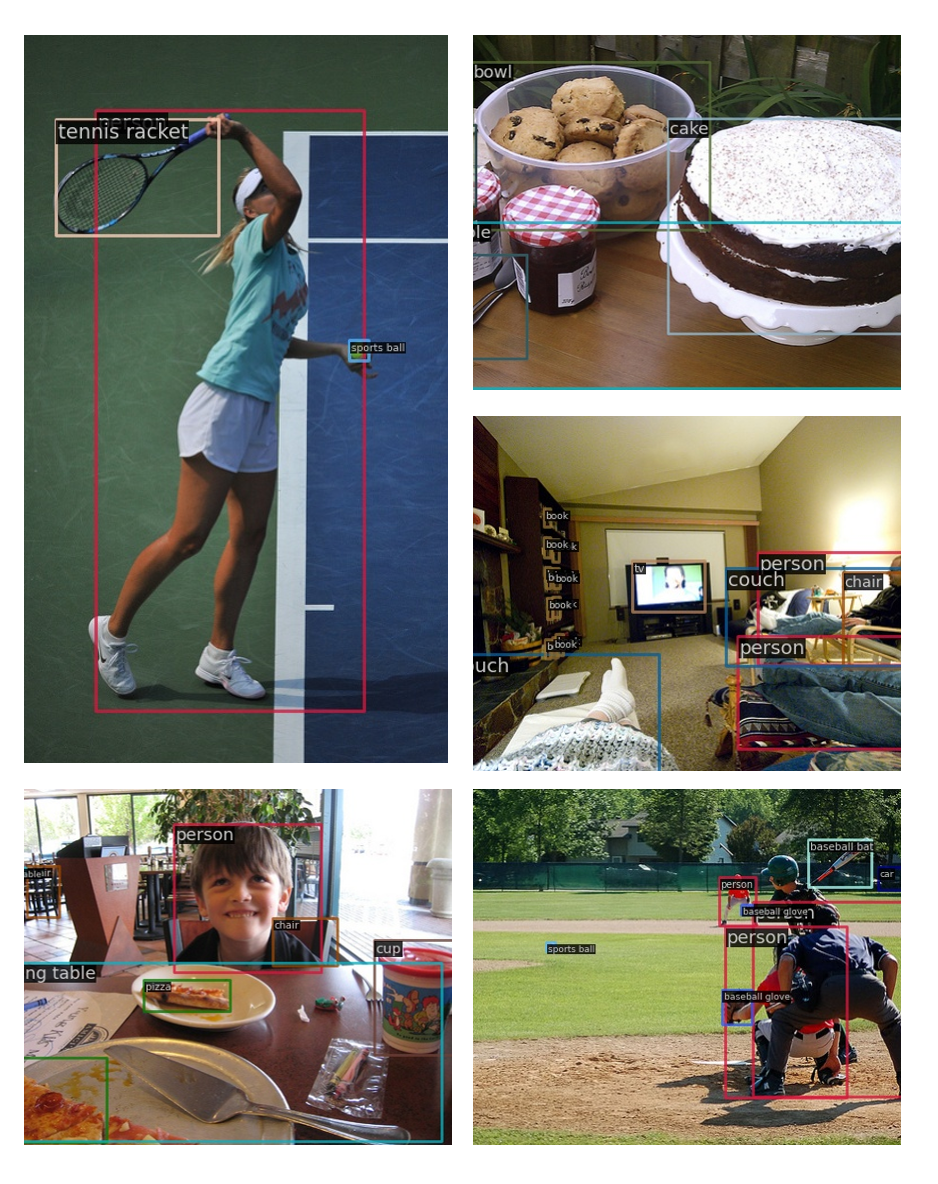}
\end{center}
  \caption{Detection results on MS COCO dataset.}
\label{fig:det}
\end{figure}
\begin{table*}[ht!]
\begin{center}
\begin{tabular}{|l|l|c|c|cccccc|}
\hline
\multirow{3}{*}{Model} & \multirow{3}{*}{Backbone} & \multirow{3}{*}{\begin{tabular}[c]{@{}c@{}}Params \\ (M)\end{tabular}} & \multirow{3}{*}{\begin{tabular}[c]{@{}c@{}}mAP\\ @320\end{tabular}} & \multicolumn{6}{c|}{Image Resolution} \\ \cline{5-10} 
 &  &  &  & \multicolumn{2}{c|}{$320^2$} & \multicolumn{2}{c|}{$640^2$} & \multicolumn{2}{c|}{$1024^2$} \\ \cline{5-10} 
 &  &  &  & FPS & \multicolumn{1}{c|}{FLOPs} & FPS & \multicolumn{1}{c|}{FLOPs} & FPS & FLOPs \\ \hline
YOLOS & DeiT & 6.5 & 25.0 & 65 & \multicolumn{1}{c|}{2.7} & 44 & \multicolumn{1}{c|}{9.3} & 10 & 22.9 \\
CenterNet & ResNet-18 & 14.4 & 25.3 & 73 & \multicolumn{1}{c|}{4.9} & 49 & \multicolumn{1}{c|}{18.2} & 28 & 46.7 \\
YOLOX & Darknet & 5.1 & 28.5 & 48 & \multicolumn{1}{c|}{1.9} & 41 & \multicolumn{1}{c|}{7.6} & 37 & 19.5 \\ \hline
\multirow{3}{*}{YOLOv3} & MobileNetV2 & 3.7 & 22.7 & 60 & \multicolumn{1}{c|}{1.7} & 55 & \multicolumn{1}{c|}{6.8} & 51 & 17.3 \\
 & MobileViT & 6.4 & 28.2 & 42 & \multicolumn{1}{c|}{3.9} & 39 & \multicolumn{1}{c|}{15.7} & 24 & 40.2 \\
 & \textbf{XFormer} & 6.3 & \textbf{33.2} & 36 & \multicolumn{1}{c|}{3.8} & \multicolumn{1}{c}{33} & \multicolumn{1}{c|}{15.0} & 27 & 38.4 \\ \hline
\end{tabular}
\end{center}
\caption{\label{tab:det}Object detection results on MS COCO \textit{val} set. Models are implemented in \textit{mmdetection} \cite{mmdet} for fair comparison. FPS is tested on a single Titan RTX 24GB GPU.}
\label{tab:det}
\end{table*}
\begin{table*}[!ht]
\begin{center}
\begin{tabular}{llcccc}
\hline
Method & Backbone & Params (M) & FLOPs (G) & mIoU (\%) & FPS \\ \hline
\multirow{2}{*}{FCN \cite{fcn}} & HRNet \cite{hrnet} & 3.9 & 38.6 & 76.3 & 23.7\\
 & MobileNetV2 & 9.7 & 158 & 64.6 & 14.2\\ \hline
PSPNet \cite{psp} & \multirow{2}{*}{MobileNetV2} & 13.6 & 211 & 70.2 & 11.2\\
DeepLabV3+ \cite{deeplabv3+}&  & 15.2 & 273 & 75.2 & 8.4\\ \hline
\multirow{4}{*}{SegFormer} & LVT \cite{lvt} & 3.9 & 39.3 & 74.9 & 9.7\\
 & MobileViT & 5.4 & 50.5 & 75.6 & 7.2\\
 & MiT-B0 & 3.8 & 25.7 & 76.5 & 12.8\\
 & \textbf{XFormer} & 5.3 & 48.7 & \textbf{78.5} & \textbf{15.3}
\\ \hline
\end{tabular}
\end{center}
\caption{\label{tab:seg}Semantic segmentation results on Cityscapes \textit{val} set. Models are implemented in \textit{mmsegmentation} \cite{mmseg} for fair comparison. FPS is tested on a single Titan RTX 24GB GPU with input of $2048 \times 1024$.}
\label{tab:seg}
\end{table*}
\subsection{Object Detection}
\noindent \textbf{Dataset.} 
For object detection task, we use MS COCO \cite{COCO} dataset, a large-scale object detection, segmentation, and captioning dataset, which contains 118K training and 5K validation images with 80 classes of objects.\\[6pt]
\noindent \textbf{Implementation details.} 
We pretrain the encoder on the ImageNet-1K dataset and randomly initialize the detector. We choose YOLOv3 \cite{yolov3} as our detector network, which is widely used for mobile devices. We use random cropping, resizing, random flipping and photo-metric distortion for data augmentation. 
We use SGD optimizer to train models for 300 epochs, with initial learning rate of 0.0001 and linear step scheduler during 24th and 28th epoch. Weight decay is 0.0005 and maximum gradient norm is set to 35. During inference, we adopt multi scale flipping augmentation test using scale of $320 \times 320$. We report object detection performance using mean Average Precision (mAP) metrics and inference speed (FPS). \\[6pt]
\noindent \textbf{Results.}
Table \ref{tab:det} shows that within YOLOv3 framework, XFormer consistently outperforms previous mobile methods for object detection. The performance of YOLOv3 is vastly improved by 46\% ($22.7 \to 33.2$ AP) from MobileNetV2 baseline, and is more accurate than state-of-art detection models, e.g., CenterNet \cite{centernet} ($+7.9$ AP), YOLOX \cite{yolox} ($+4.7$ AP), while maintaining relatively light-weight (6.4M parameter, 3.8G FLOPs). In terms of inference speed, especially for high resolution input, XFormer computes faster than ViT-based detection networks, such as YOLOS \cite{yolos} ($+17$ FPS) and MobileViT \cite{mobilevit} ($+3$ FPS), which shows our method is indeed more computationally efficient thanks to our XFA module. This demonstrates that our proposed XFormer model can learn efficient representation for object detection task. Qualitative results can be seen in Fig. \ref{fig:det}.

\begin{figure*}[!ht]
\begin{center}
  \includegraphics[width=1.0\linewidth]{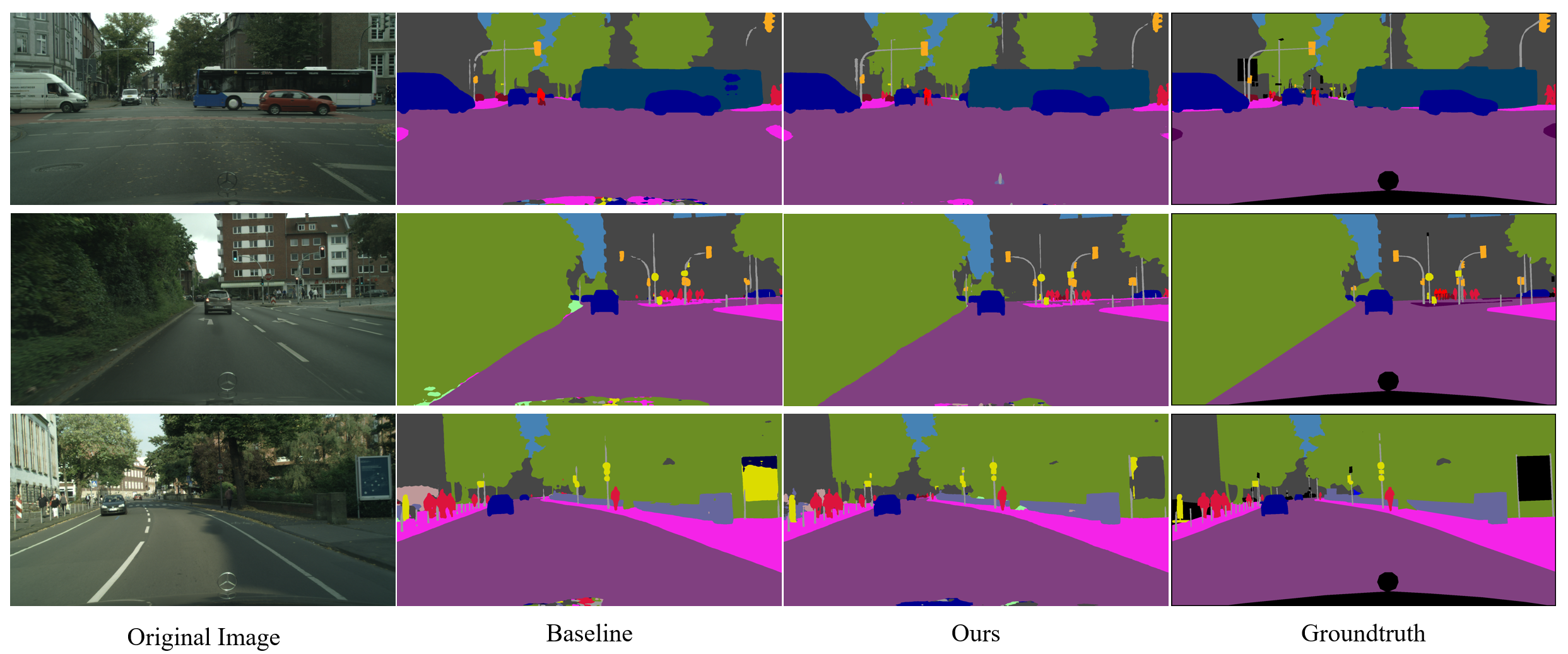}
\end{center}
  \caption{Segmentation results on Cityscapes dataset. From left to right are original images, baseline segmentation results (FCN w/ MobileNetV2), segmentation results of our method and groundtruth masks.}
\label{fig:semantic}
\end{figure*}

\subsection{Semantic Segmentation}
\noindent \textbf{Dataset.} 
We use Cityscapes \cite{Cityscapes} dataset to train and evaluate our method for semantic segmentation task. Cityscapes is a challenging semantic segmentation dataset with 5,000 finely annotated high resolution driving images and 19 categories in total.\\[6pt]
\noindent \textbf{Implementation details.} 
We pretrain encoder on the ImageNet-1K dataset and choose the simple all-MLP decoder from SegFormer \cite{Segformer}, which consists only linear projection and upsampling operations. We use random resizing (0.5-2.0), random horizontal flipping and random cropping to size $1024 \times 1024$. We use AdamW optimizer to train models for 160k iterations, with initial learning rate of 0.00005 and polynomial scheduler with factor of 1 by default. During inference, we use a sliding window test by cropping $1024 \times 1024$ windows with strides of $768 \times 768$. We report semantic segmentation performance using mean Intersection over Union (mIoU) metrics inference speed (FPS).\\[6pt]
\noindent \textbf{Results.}
Table \ref{tab:seg} shows that XFormer achieves competitive results on Cityscapes dataset against state-of-art mobile methods for semantic segmentation. Compared with CNN-based models, our method delivers better performance than MobileNetV2, without the need for heavy decoders. Furthermore, XFormer improves SegFormer with MobileViT by 3.8\% ($75.6 \to 78.5$ mIoU), and also outperforms state-of-the-art light-weight LVT \cite{lvt} ($+3.6$ mIoU) and SegFormer \cite{Segformer} ($+2.0$ mIoU). Above all, XFormer achieves faster inference speed against previous ViT backbones, validating our model's efficiency in processing high resolution inputs. This suggests that our proposed XFormer model can serve as an effective backbone for semantic segmentation task. Qualitative results can be seen in Fig. \ref{fig:semantic}.
\section{Conclusion and Future Work}
\noindent In this paper, we present cross feature attention (XFA), which reduces computation cost for transformers by handling attention via feature dimension and produces only linear complexity. We also propose XFormer, a light-weight CNN-ViT hybrid model, building on cross feature attention (XFA) transformers and efficient CNN blocks. It leverages both the local processing ability of efficient convolutional layer and the global encoding ability of light-weight transformers. Experimental results on various datasets and tasks prove our efficient design can not only improve performance against previous methods, but also maintain low computation cost and high efficiency.
In our future work we plan to explore other methods such as NAS \cite{sss} to continue optimizing our model in terms of FLOPs and inference speed. Self-supervised pretraining is another direction to investigate to help our model generalize better on unseen datasets.
\section{Acknowledgements}
\noindent We want to thank Zhuoxin Gan and Weixin Mao for their helpful suggestions without which this paper would not be possible.

{\small
\bibliographystyle{ieee_fullname}
\bibliography{refs}

\begin{thebibliography}{10}\itemsep=-1pt

\bibitem{ofa}
Han Cai, Chuang Gan, and Song Han.
\newblock Once for all: Train one network and specialize it for efficient
  deployment.
\newblock {\em ArXiv}, abs/1908.09791, 2020.

\bibitem{proxylessnas}
Han Cai, Ligeng Zhu, and Song Han.
\newblock Proxylessnas: Direct neural architecture search on target task and
  hardware.
\newblock {\em ArXiv}, abs/1812.00332, 2019.

\bibitem{crossvit}
Chun-Fu Chen, Quanfu Fan, and Rameswar Panda.
\newblock Crossvit: Cross-attention multi-scale vision transformer for image
  classification.
\newblock {\em 2021 IEEE/CVF International Conference on Computer Vision
  (ICCV)}, pages 347--356, 2021.

\bibitem{addernet1}
Hanting Chen, Yunhe Wang, Chunjing Xu, Boxin Shi, Chao Xu, Qi Tian, and Chang
  Xu.
\newblock Addernet: Do we really need multiplications in deep learning?
\newblock {\em 2020 IEEE/CVF Conference on Computer Vision and Pattern
  Recognition (CVPR)}, pages 1465--1474, 2020.

\bibitem{mmdet}
Kai Chen, Jiaqi Wang, Jiangmiao Pang, Yuhang Cao, Yu Xiong, Xiaoxiao Li,
  Shuyang Sun, Wansen Feng, Ziwei Liu, Jiarui Xu, Zheng Zhang, Dazhi Cheng,
  Chenchen Zhu, Tianheng Cheng, Qijie Zhao, Buyu Li, Xin Lu, Rui Zhu, Yue Wu,
  Jifeng Dai, Jingdong Wang, Jianping Shi, Wanli Ouyang, Chen~Change Loy, and
  Dahua Lin.
\newblock {MMDetection}: Open mmlab detection toolbox and benchmark.
\newblock {\em arXiv preprint arXiv:1906.07155}, 2019.

\bibitem{deeplabv3+}
Liang-Chieh Chen, Yukun Zhu, George Papandreou, Florian Schroff, and Hartwig
  Adam.
\newblock Encoder-decoder with atrous separable convolution for semantic image
  segmentation.
\newblock In {\em ECCV}, 2018.

\bibitem{sss}
Minghao Chen, Kan Wu, Bolin Ni, Houwen Peng, Bei Liu, Jianlong Fu, Hongyang
  Chao, and Haibin Ling.
\newblock Searching the search space of vision transformer.
\newblock In {\em NeurIPS}, 2021.

\bibitem{mobileformer}
Yinpeng Chen, Xiyang Dai, Dongdong Chen, Mengchen Liu, Xiaoyi Dong, Lu Yuan,
  and Zicheng Liu.
\newblock Mobile-former: Bridging mobilenet and transformer.
\newblock {\em ArXiv}, abs/2108.05895, 2021.

\bibitem{sparsevit}
Rewon Child, Scott Gray, Alec Radford, and Ilya Sutskever.
\newblock Generating long sequences with sparse transformers.
\newblock {\em ArXiv}, abs/1904.10509, 2019.

\bibitem{gpu2}
Dan~C. Ciresan, Ueli Meier, and J{\"u}rgen Schmidhuber.
\newblock Multi-column deep neural networks for image classification.
\newblock {\em 2012 IEEE Conference on Computer Vision and Pattern
  Recognition}, pages 3642--3649, 2012.

\bibitem{mmseg}
MMSegmentation Contributors.
\newblock {MMSegmentation}: Openmmlab semantic segmentation toolbox and
  benchmark.
\newblock \url{https://github.com/open-mmlab/mmsegmentation}, 2020.

\bibitem{Cityscapes}
Marius Cordts, Mohamed Omran, Sebastian Ramos, Timo Rehfeld, Markus Enzweiler,
  Rodrigo Benenson, Uwe Franke, Stefan Roth, and Bernt Schiele.
\newblock The cityscapes dataset for semantic urban scene understanding.
\newblock {\em 2016 IEEE Conference on Computer Vision and Pattern Recognition
  (CVPR)}, pages 3213--3223, 2016.

\bibitem{randaug}
Ekin~Dogus Cubuk, Barret Zoph, Jonathon Shlens, and Quoc~V. Le.
\newblock Randaugment: Practical automated data augmentation with a reduced
  search space.
\newblock {\em 2020 IEEE/CVF Conference on Computer Vision and Pattern
  Recognition Workshops (CVPRW)}, pages 3008--3017, 2020.

\bibitem{convvit}
St{\'e}phane d'Ascoli, Hugo Touvron, Matthew~L. Leavitt, Ari~S. Morcos, Giulio
  Biroli, and Levent Sagun.
\newblock Convit: Improving vision transformers with soft convolutional
  inductive biases.
\newblock In {\em ICML}, 2021.

\bibitem{ImageNet}
Jia Deng, Wei Dong, Richard Socher, Li-Jia Li, K. Li, and Li Fei-Fei.
\newblock Imagenet: A large-scale hierarchical image database.
\newblock In {\em CVPR}, 2009.

\bibitem{prun1}
Emily~L. Denton, Wojciech Zaremba, Joan Bruna, Yann LeCun, and Rob Fergus.
\newblock Exploiting linear structure within convolutional networks for
  efficient evaluation.
\newblock {\em ArXiv}, abs/1404.0736, 2014.

\bibitem{vit}
Alexey Dosovitskiy, Lucas Beyer, Alexander Kolesnikov, Dirk Weissenborn,
  Xiaohua Zhai, Thomas Unterthiner, Mostafa Dehghani, Matthias Minderer, Georg
  Heigold, Sylvain Gelly, Jakob Uszkoreit, and Neil Houlsby.
\newblock An image is worth 16x16 words: Transformers for image recognition at
  scale.
\newblock {\em ArXiv}, abs/2010.11929, 2021.

\bibitem{centernet}
Kaiwen Duan, Song Bai, Lingxi Xie, Honggang Qi, Qingming Huang, and Qi Tian.
\newblock Centernet: Keypoint triplets for object detection.
\newblock {\em 2019 IEEE/CVF International Conference on Computer Vision
  (ICCV)}, pages 6568--6577, 2019.

\bibitem{xcit}
Alaaeldin El-Nouby, Hugo Touvron, Mathilde Caron, Piotr Bojanowski, Matthijs
  Douze, Armand Joulin, Ivan Laptev, Natalia Neverova, Gabriel Synnaeve, Jakob
  Verbeek, and Herv{\'e} J{\'e}gou.
\newblock Xcit: Cross-covariance image transformers.
\newblock {\em ArXiv}, abs/2106.09681, 2021.

\bibitem{SiLU}
Stefan Elfwing, Eiji Uchibe, and Kenji Doya.
\newblock Sigmoid-weighted linear units for neural network function
  approximation in reinforcement learning.
\newblock {\em Neural networks : the official journal of the International
  Neural Network Society}, 107:3--11, 2018.

\bibitem{yolos}
Yuxin Fang, Bencheng Liao, Xinggang Wang, Jiemin Fang, Jiyang Qi, Rui Wu,
  Jianwei Niu, and Wenyu Liu.
\newblock You only look at one sequence: Rethinking transformer in vision
  through object detection.
\newblock In {\em NeurIPS}, 2021.

\bibitem{yolox}
Zheng Ge, Songtao Liu, Feng Wang, Zeming Li, and Jian Sun.
\newblock Yolox: Exceeding yolo series in 2021.
\newblock {\em ArXiv}, abs/2107.08430, 2021.

\bibitem{prun2}
Song Han, Huizi Mao, and William~J. Dally.
\newblock Deep compression: Compressing deep neural network with pruning,
  trained quantization and huffman coding.
\newblock {\em arXiv: Computer Vision and Pattern Recognition}, 2016.

\bibitem{resnet}
Kaiming He, X. Zhang, Shaoqing Ren, and Jian Sun.
\newblock Deep residual learning for image recognition.
\newblock {\em 2016 IEEE Conference on Computer Vision and Pattern Recognition
  (CVPR)}, pages 770--778, 2016.

\bibitem{gelu}
Dan Hendrycks and Kevin Gimpel.
\newblock Gaussian error linear units (gelus).
\newblock {\em arXiv: Learning}, 2016.

\bibitem{kd1}
Geoffrey~E. Hinton, Oriol Vinyals, and Jeffrey Dean.
\newblock Distilling the knowledge in a neural network.
\newblock {\em ArXiv}, abs/1503.02531, 2015.

\bibitem{mbv3}
Andrew~G. Howard, Mark Sandler, Grace Chu, Liang-Chieh Chen, Bo Chen, Mingxing
  Tan, Weijun Wang, Yukun Zhu, Ruoming Pang, Vijay Vasudevan, Quoc~V. Le, and
  Hartwig Adam.
\newblock Searching for mobilenetv3.
\newblock {\em 2019 IEEE/CVF International Conference on Computer Vision
  (ICCV)}, pages 1314--1324, 2019.

\bibitem{mbv}
Andrew~G. Howard, Menglong Zhu, Bo Chen, Dmitry Kalenichenko, Weijun Wang,
  Tobias Weyand, Marco Andreetto, and Hartwig Adam.
\newblock Mobilenets: Efficient convolutional neural networks for mobile vision
  applications.
\newblock {\em ArXiv}, abs/1704.04861, 2017.

\bibitem{prun3}
Hengyuan Hu, Rui Peng, Yu-Wing Tai, and Chi-Keung Tang.
\newblock Network trimming: A data-driven neuron pruning approach towards
  efficient deep architectures.
\newblock {\em ArXiv}, abs/1607.03250, 2016.

\bibitem{se}
Jie Hu, Li Shen, Samuel Albanie, Gang Sun, and Enhua Wu.
\newblock Squeeze-and-excitation networks.
\newblock {\em IEEE Transactions on Pattern Analysis and Machine Intelligence},
  42:2011--2023, 2020.

\bibitem{densenet}
Gao Huang, Zhuang Liu, and Kilian~Q. Weinberger.
\newblock Densely connected convolutional networks.
\newblock {\em 2017 IEEE Conference on Computer Vision and Pattern Recognition
  (CVPR)}, pages 2261--2269, 2017.

\bibitem{reformer}
Nikita Kitaev, Lukasz Kaiser, and Anselm Levskaya.
\newblock Reformer: The efficient transformer.
\newblock {\em ArXiv}, abs/2001.04451, 2020.

\bibitem{alexnet}
Alex Krizhevsky, Ilya Sutskever, and Geoffrey~E. Hinton.
\newblock Imagenet classification with deep convolutional neural networks.
\newblock {\em Communications of the ACM}, 60:84 -- 90, 2012.

\bibitem{lenet}
Yann LeCun, L{\'e}on Bottou, Yoshua Bengio, and Patrick Haffner.
\newblock Gradient-based learning applied to document recognition.
\newblock {\em Proc. IEEE}, 86:2278--2324, 1998.

\bibitem{vitdet}
Yanghao Li, Hanzi Mao, Ross~B. Girshick, and Kaiming He.
\newblock Exploring plain vision transformer backbones for object detection.
\newblock {\em ArXiv}, abs/2203.16527, 2022.

\bibitem{COCO}
Tsung-Yi Lin, Michael Maire, Serge~J. Belongie, James Hays, Pietro Perona, Deva
  Ramanan, Piotr Doll{\'a}r, and C.~Lawrence Zitnick.
\newblock Microsoft coco: Common objects in context.
\newblock In {\em ECCV}, 2014.

\bibitem{darts}
Hanxiao Liu, Karen Simonyan, and Yiming Yang.
\newblock Darts: Differentiable architecture search.
\newblock {\em ArXiv}, abs/1806.09055, 2019.

\bibitem{swin}
Ze Liu, Yutong Lin, Yue Cao, Han Hu, Yixuan Wei, Zheng Zhang, Stephen Lin, and
  Baining Guo.
\newblock Swin transformer: Hierarchical vision transformer using shifted
  windows.
\newblock {\em 2021 IEEE/CVF International Conference on Computer Vision
  (ICCV)}, pages 9992--10002, 2021.

\bibitem{fcn}
Jonathan Long, Evan Shelhamer, and Trevor Darrell.
\newblock Fully convolutional networks for semantic segmentation.
\newblock {\em 2015 IEEE Conference on Computer Vision and Pattern Recognition
  (CVPR)}, pages 3431--3440, 2015.

\bibitem{adamw}
Ilya Loshchilov and Frank Hutter.
\newblock Decoupled weight decay regularization.
\newblock In {\em ICLR}, 2019.

\bibitem{shufflev2}
Ningning Ma, Xiangyu Zhang, Haitao Zheng, and Jian Sun.
\newblock Shufflenet v2: Practical guidelines for efficient cnn architecture
  design.
\newblock In {\em ECCV}, 2018.

\bibitem{mobilevit}
Sachin Mehta and Mohammad Rastegari.
\newblock Mobilevit: Light-weight, general-purpose, and mobile-friendly vision
  transformer.
\newblock {\em ArXiv}, abs/2110.02178, 2021.

\bibitem{gpu1}
Rajat Raina, Anand Madhavan, and A. Ng.
\newblock Large-scale deep unsupervised learning using graphics processors.
\newblock In {\em ICML '09}, 2009.

\bibitem{yolov3}
Joseph Redmon and Ali Farhadi.
\newblock Yolov3: An incremental improvement.
\newblock {\em ArXiv}, abs/1804.02767, 2018.

\bibitem{kd2}
Adriana Romero, Nicolas Ballas, Samira~Ebrahimi Kahou, Antoine Chassang, Carlo
  Gatta, and Yoshua Bengio.
\newblock Fitnets: Hints for thin deep nets.
\newblock {\em CoRR}, abs/1412.6550, 2015.

\bibitem{mbv2}
Mark Sandler, Andrew~G. Howard, Menglong Zhu, Andrey Zhmoginov, and Liang-Chieh
  Chen.
\newblock Mobilenetv2: Inverted residuals and linear bottlenecks.
\newblock {\em 2018 IEEE/CVF Conference on Computer Vision and Pattern
  Recognition}, pages 4510--4520, 2018.

\bibitem{vgg}
Karen Simonyan and Andrew Zisserman.
\newblock Very deep convolutional networks for large-scale image recognition.
\newblock {\em CoRR}, abs/1409.1556, 2015.

\bibitem{segmentor}
Robin Strudel, Ricardo~Garcia Pinel, Ivan Laptev, and Cordelia Schmid.
\newblock Segmenter: Transformer for semantic segmentation.
\newblock {\em 2021 IEEE/CVF International Conference on Computer Vision
  (ICCV)}, pages 7242--7252, 2021.

\bibitem{jft}
Chen Sun, Abhinav Shrivastava, Saurabh Singh, and Abhinav~Kumar Gupta.
\newblock Revisiting unreasonable effectiveness of data in deep learning era.
\newblock {\em 2017 IEEE International Conference on Computer Vision (ICCV)},
  pages 843--852, 2017.

\bibitem{efficientnet}
Mingxing Tan and Quoc~V. Le.
\newblock Efficientnet: Rethinking model scaling for convolutional neural
  networks.
\newblock {\em ArXiv}, abs/1905.11946, 2019.

\bibitem{deit}
Hugo Touvron, Matthieu Cord, Matthijs Douze, Francisco Massa, Alexandre
  Sablayrolles, and Herv'e J'egou.
\newblock Training data-efficient image transformers \& distillation through
  attention.
\newblock In {\em ICML}, 2021.

\bibitem{attention}
Ashish Vaswani, Noam~M. Shazeer, Niki Parmar, Jakob Uszkoreit, Llion Jones,
  Aidan~N. Gomez, Lukasz Kaiser, and Illia Polosukhin.
\newblock Attention is all you need.
\newblock {\em ArXiv}, abs/1706.03762, 2017.

\bibitem{hrnet}
Jingdong Wang, Ke Sun, Tianheng Cheng, Borui Jiang, Chaorui Deng, Yang Zhao,
  Dong Liu, Yadong Mu, Mingkui Tan, Xinggang Wang, Wenyu Liu, and Bin Xiao.
\newblock Deep high-resolution representation learning for visual recognition.
\newblock {\em IEEE Transactions on Pattern Analysis and Machine Intelligence},
  43:3349--3364, 2021.

\bibitem{linformer}
Sinong Wang, Belinda~Z. Li, Madian Khabsa, Han Fang, and Hao Ma.
\newblock Linformer: Self-attention with linear complexity.
\newblock {\em ArXiv}, abs/2006.04768, 2020.

\bibitem{addernet2}
Yunhe Wang, Mingqiang Huang, Kai Han, Hanting Chen, Wei Zhang, Chunjing Xu, and
  Dacheng Tao.
\newblock Addernet and its minimalist hardware design for energy-efficient
  artificial intelligence.
\newblock {\em ArXiv}, abs/2101.10015, 2021.

\bibitem{dynamicvit}
Yulin Wang, Rui Huang, Shiji Song, Zeyi Huang, and Gao Huang.
\newblock Not all images are worth 16x16 words: Dynamic transformers for
  efficient image recognition.
\newblock In {\em NeurIPS}, 2021.

\bibitem{vitc}
Tete Xiao, Mannat Singh, Eric Mintun, Trevor Darrell, Piotr Doll{\'a}r, and
  Ross~B. Girshick.
\newblock Early convolutions help transformers see better.
\newblock In {\em NeurIPS}, 2021.

\bibitem{Segformer}
Enze Xie, Wenhai Wang, Zhiding Yu, Anima Anandkumar, Jos{\'e}~Manuel
  {\'A}lvarez, and Ping Luo.
\newblock Segformer: Simple and efficient design for semantic segmentation with
  transformers.
\newblock In {\em NeurIPS}, 2021.

\bibitem{lvt}
Chenglin Yang, Yilin Wang, Jianming Zhang, He Zhang, Zijun Wei, Zhe~L. Lin, and
  Alan~Loddon Yuille.
\newblock Lite vision transformer with enhanced self-attention.
\newblock {\em ArXiv}, abs/2112.10809, 2021.

\bibitem{kd3}
Shan You, Chang Xu, Chao Xu, and Dacheng Tao.
\newblock Learning from multiple teacher networks.
\newblock {\em Proceedings of the 23rd ACM SIGKDD International Conference on
  Knowledge Discovery and Data Mining}, 2017.

\bibitem{mixup}
Hongyi Zhang, Moustapha Ciss{\'e}, Yann Dauphin, and David Lopez-Paz.
\newblock mixup: Beyond empirical risk minimization.
\newblock {\em ArXiv}, abs/1710.09412, 2018.

\bibitem{shufflev1}
Xiangyu Zhang, Xinyu Zhou, Mengxiao Lin, and Jian Sun.
\newblock Shufflenet: An extremely efficient convolutional neural network for
  mobile devices.
\newblock {\em 2018 IEEE/CVF Conference on Computer Vision and Pattern
  Recognition}, pages 6848--6856, 2018.

\bibitem{psp}
Hengshuang Zhao, Jianping Shi, Xiaojuan Qi, Xiaogang Wang, and Jiaya Jia.
\newblock Pyramid scene parsing network.
\newblock {\em 2017 IEEE Conference on Computer Vision and Pattern Recognition
  (CVPR)}, pages 6230--6239, 2017.

\bibitem{setr}
Sixiao Zheng, Jiachen Lu, Hengshuang Zhao, Xiatian Zhu, Zekun Luo, Yabiao Wang,
  Yanwei Fu, Jianfeng Feng, Tao Xiang, Philip H.~S. Torr, and Li Zhang.
\newblock Rethinking semantic segmentation from a sequence-to-sequence
  perspective with transformers.
\newblock {\em 2021 IEEE/CVF Conference on Computer Vision and Pattern
  Recognition (CVPR)}, pages 6877--6886, 2021.

\bibitem{randomerasing}
Zhun Zhong, Liang Zheng, Guoliang Kang, Shaozi Li, and Yi Yang.
\newblock Random erasing data augmentation.
\newblock In {\em AAAI}, 2020.

\end{thebibliography}
}

\end{document}